\documentclass[10pt,twocolumn,letterpaper]{article}

\usepackage{iccv}
\usepackage{times}
\usepackage{epsfig}
\usepackage{graphicx}
\usepackage{amsmath}
\usepackage{amssymb}
\usepackage{cuted}
\usepackage{capt-of}
\usepackage{adjustbox}
\usepackage{comment}
\usepackage[accsupp]{axessibility}  

\newcommand{\ad}[1]{\textcolor{black}{#1}}
\newcommand{\aj}[1]{\textcolor{black}{#1}}
\newcommand{\CeT}[1]{\textcolor{black}{#1}}

\newcommand{\fv}[1]{\textcolor{black}{#1}}

\newcommand{\md}[1]{\textcolor{yellow}{}}

\usepackage[pagebackref=true,breaklinks=true,letterpaper=true,colorlinks,bookmarks=false]{hyperref}

\iccvfinalcopy 

\def\iccvPaperID{9084} 

\ificcvfinal\fi

\begin{document}
\title{Towards High Fidelity Monocular Face Reconstruction with Rich Reflectance using Self-supervised Learning and Ray Tracing}

\author
{\parbox{\textwidth}{\centering Abdallah Dib$^{1}$\thanks{Equal contribution}\;\;
                                C\'edric Th\'ebault$^{1*}$\;\;
                                Junghyun Ahn$^{1*}$\;\;
                                Philippe-Henri Gosselin$^{1}$\;\;
                                \\
                                Christian Theobalt$^{2}$\;\;
                                Louis Chevallier$^{1}$
        }
        \\
        \\
{\parbox{\textwidth}{\centering $^1$InterDigital R\&I\;\;\;\;\;
                                $^2$Max-Planck-Institute for Informatics
       }
}
\vspace{-20px}
}
\maketitle
\ificcvfinal\thispagestyle{empty}\fi
\begin{strip}\centering
\includegraphics[width=\textwidth]{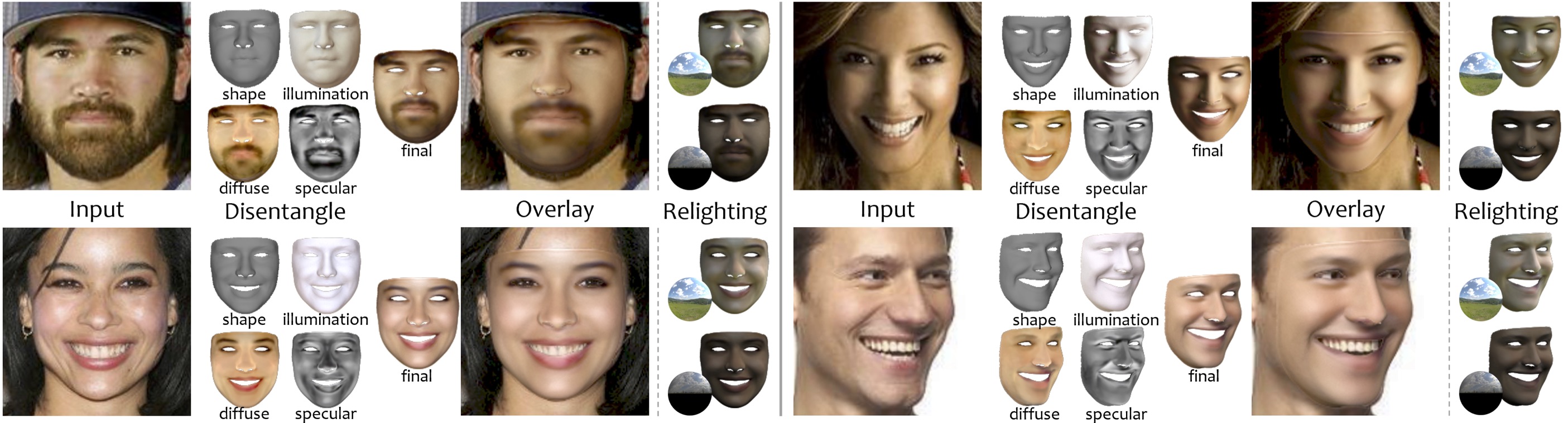}
\captionof{figure}{
A deep neural network trained in a self-supervised manner with a differentiable ray tracer estimates a complete set of facial attributes -- 3D head pose, geometry, personalized albedo (diffuse and specular) from unconstrained monocular image. Reconstruction based on these attributes enables a variety of applications, such as relighting.
\label{fig:teaser}}
\end{strip}
\begin{abstract}
\vspace{-10px}
Robust face reconstruction from monocular image in general lighting conditions is challenging. Methods combining deep neural network encoders with differentiable rendering have opened up the path for very fast monocular reconstruction of geometry, lighting and reflectance. They can also be trained in self-supervised manner for increased robustness and better generalization. However, their differentiable rasterization-based image formation models, as well as underlying scene parameterization, limit them to Lambertian face reflectance and to poor shape details. 
More recently, ray tracing was introduced for monocular face reconstruction within a classic optimization-based framework and enables state-of-the art results. However, optimization-based approaches are inherently slow and lack robustness. In this paper, we build our work on the aforementioned approaches and propose a new method that greatly improves reconstruction quality and robustness in general scenes. We achieve this by combining a CNN encoder with a differentiable ray tracer, which enables us to base the reconstruction on much more advanced personalized diffuse and specular albedos, a more sophisticated illumination model and a plausible representation of self-shadows. This enables to take a big leap forward in reconstruction quality of shape, appearance and lighting even in scenes with difficult illumination.
With consistent face attributes reconstruction, our method leads to practical applications such as relighting and self-shadows removal. Compared to state-of-the-art methods, our results show improved accuracy and validity of the approach.


\end{abstract}

\vspace{-10px}
\section{Introduction}
\label{sec:intro}
Fast and accurate image-based face reconstruction has many applications in several domains including rig based realistic videoconferencing, interactive AR/VR experiences and special effects for professionals like facial attribute manipulation/transfer or relighting.
Also, supporting unconstrained pose and in-the-wild capture conditions without specific hardware, such as multi-view setup, allows for enhanced flexibility and extended applicability. 
However, captured images reflect the complex interaction between light and faces including shadows and specularities, which poses a real challenge for face reconstruction.
Speed is also a key factor for interactive scenario and other real-time use cases.
Great multi-view approaches exist (\cite{beeler2011high, wu2011shading, valgaerts2012lightweight,ghosh2011multiview, gotardo2018practical}), but they may not be easily applied in many applications such as VR or movies / special effects.
Significant progress has been made on monocular face reconstruction where most methods resort to some form of parametric prior; high-quality analysis-by-synthesis monocular optimization methods exist (\cite{ Garrido2013,suwajanakorn2014total,Garrido:2016}), but besides being rather slow, they would fail for difficult head poses and lighting conditions. 
\CeT{More recently, and to improve this robustness against lighting conditions, \cite{dib2021practical} introduced ray tracing for face reconstruction within an optimization-based framework. But the quality of their reconstruction remains sensitive to the landmarks used for initialization.
}

Real-time analysis-by-synthesis approaches have also been presented, however they often sacrifice reconstruction details. To increase reconstruction efficiency, CNN based approaches (\cite{tewari17MoFA, tewari18fml, tran2018nonlinear, tran2019towards}) that directly regress 3D reconstruction parameters from images have been investigated. 
To overcome the challenge of creating large amounts of labeled data, while enabling reconstruction on the basis of meaningful scene parameters, methods combining CNNs with differentiable image formation models trained in a self-supervised way have been presented (\cite{tewari17MoFA, tewari18fml, tewari2019fml}). 
They enable reconstruction performance in the range of milliseconds, and can be applied to more general scenes and subjects (\cite{tewari18fml, tewari2019fml}).
However, even the best of these highly efficient monocular reconstruction methods fall short of the quality and robustness requirements expected in professional visual effects (VFX) pipelines.
They rely on simple parametric diffuse reflectance models based on low-frequency spherical harmonics (SH) illumination model, whereas more detailed models would be needed 
to reconstruct at the level of quality that is typically required. The inability to model self-shadows is also a prime reason for their instability under challenging scene conditions.

To overcome these limitations, we present a new approach that is the first to jointly provide the following capabilities: it enables monocular reconstruction of detailed face geometry, spatially varying face reflectance and complex scene illumination at very high speed on the basis of semantically meaningful scene parameters.
To achieve this, our model resorts to a parametric face and scene model that represents geometry using 3DMM statistical model, illumination as high-order spherical harmonics and reflectance model with diffuse and specular components. Our new CNN-based approach can be trained in a self-supervised way on unlabeled image data. 
It features a CNN encoder projecting the input image into the parametric scene representation. We also use an end-to-end differentiable ray tracing image formation model which, in contrast to earlier rasterization-based models, provides a more accurate light-geometry interaction and is able to synthesize images with complex illumination accounting for self-shadows. To the best of our knowledge, this is the first time a differentiable ray tracer is used for deep-based face reconstruction in an inverse-rendering setup. \ad{While ray tracing enables \cite{dib2021practical} to improve the state of the art, their method is based on costly and slow iterative optimization and the final reconstruction remains sensitive to the quality of the landmarks.
Our method overcomes these limitations: it absorbs the complexity of ray tracing at training time, achieves robust and competitive results with near real-time performance, and, being completely independent of landmarks at test time, is more suitable for in-the-wild conditions.
Finally, with an appropriate training strategy, our method is the first self-supervised method to achieve robust face reconstruction in challenging lighting conditions and captures person-specific shadow-free albedo details (such as facial hair or makeup) not restricted by the 3DMM space. Our rich and consistent facial attributes reconstruction naturally allows various types of applications such as relighting, light/albedo edit and transfer. Our comparison with recent state-of-the-art methods shows improved robustness, accuracy and versatility of our method.} 

\section{Related works}
\label{sec:sota}
While methods such as \cite{wu2011shading, beeler2011high, valgaerts2012lightweight, ghosh2011multiview,  gotardo2018practical} produce high-quality face reconstruction from multi-view camera and/or from multi-light illumination setup, they are not applicable for in-the-wild images. We focus therefore on monocular image-based face reconstruction approaches that do not require any external hardware setup.
\vspace{4px}
\\
\textbf{Statistical morphable model} To make the highly ill-posed problem of monocular face reconstruction tractable, statistical priors have been introduced \cite{zollhoefer2018facestar} such as 3D morphable models (3DMMs) \cite{blanz1999morphable,li2017learning,Egger20Years}.
3DMM was the main building block for most recent approaches because of its efficiency. However, 3DMM limits face reconstruction to a low-dimensional space and does not capture person specific details (such as beards and makeup), and its statistical albedo prior bakes some illumination in it. Recently, to overcome these limitations, \cite{smith2020morphable} proposes a drop-in replacement to 3DMM statistical albedo model with more sophisticated diffuse and specular albedo priors.
In this work, we base our reconstruction on the 3DMM geometry and albedo prior (diffuse and specular) of \cite{smith2020morphable} and we learn an increment, similarly to \cite{tran2019towards, dib2021practical}, on top of the albedo prior to capture more person-specific details outside of the statistical prior space.
\vspace{4px}
\\
\textbf{Optimization-based approaches} \cite{Garrido:2016, Garrido2013,suwajanakorn2014total, wu2016anatomically, andrus2020face, dib2021practical} use optimization formulation for face (geometry and diffuse albedo) reconstruction. In \cite{Garrido:2016, Garrido2013,suwajanakorn2014total, dib2021practical}, a 3DMM based statistical priors act as optimization regularizer while minimizing a photo-consistency loss function. 
While such methods work well for controlled scene conditions, often they do not generalize well for in-the-wild images scenarios and can be computationally expensive.
\vspace{4px}
\\
\textbf{Illumination and reflectance models} 
The aforementioned methods generally use low-order spherical harmonics (SH) to model light and assume Lambertian surface, while our method uses higher-order SH in order to better model light interaction with non-Lambertian surface (with diffuse and specular). While \cite{LiChenIntrinsics} extracts diffuse and specular albedos from a single image using SH illumination, they do not explicitly model self-shadows and show only results in controlled conditions. \ad{\cite{dib2021practical} introduced a novel virtual light stage to model area lights and used a Cook-Torrance BRDF (diffuse, specular and roughness) to model skin reflectance.}  
\vspace{4px}
\\
\textbf{Differentiable rendering}
A simple and efficient vertex-wise differentiable rendering is proposed in \cite{tewari17MoFA}. Two shortcomings of this approach are the simple Lambertian BRDF and its inability to handle self-shadows. To address these limitations, \cite{dib2019face, dib2021practical} introduce the use of differentiable ray tracing for face reconstruction. It can handle more complex illumination and BRDF models, and naturally accounts for self-shadows. However, in addition to being computationally expensive, differentiable ray tracing exhibits noisy gradients on the geometry edges as they are sampled by very few points (solutions, such as \cite{Loubet2019Reparameterizing,Li:2018:DMC}, exist but remain computationally expensive).
\vspace{4px}
\\
\textbf{Deep learning based approaches}
When trained on large corpus, convolution neural networks (CNNs) are well proven face deep representation decoders and encoders such as~\cite{liu2018large, karras2019style}.
Several novel approaches have been proposed (\cite{tran2018nonlinear, tran2019towards, tewari17MoFA, tewari18fml, tewari2019fml, Koizumi20Look, sengupta2018sfsnet}). \cite{tewari17MoFA} was among the first to show the use of self-supervised autoencoder-like inverse rendering based architecture to \textit{infer} semantic attributes. These unsupervised learning approaches have the potential to leverage vast amount of face images.
They are however limited by their differentiable rasterizer, relying on a low-order spherical harmonics parametrization and a pure-Lambertian skin reflectance model, which limits their performance and reconstruction quality under challenging lighting conditions. \cite{tewari17MoFA} relies on diffuse albedo obtained from 3DMM, which restricts generalization of the true diversity in face reflectance. \cite{tewari18fml} uses self-supervision to learn a corrective space to capture more person-specific albedo/geometry details outside of the 3DMM space. More recently, \cite{Chaudhuri20Personalized} learns user-specific expression blendshapes and dynamic albedo maps by predicting personalized corrections on top of a 3DMM prior. However, these methods do not separate diffuse and specular albedos, which get mixed in their final estimated reflectance. They also do not show reconstruction under challenging lighting conditions. 
Our method uses self-supervision to train a CNN and extracts personalized diffuse and specular albedos, outside of the 3DMM space. In contrast to the aforementioned methods, our method uses high-order SH for better light approximation and an image formation layer based on a differentiable ray tracing that handle advanced lighting and self-shadows. All this contributes to a robust face reconstruction even in scenes with challenging lighting conditions. 
Our loss functions and training strategy allow us to obtain personalized diffuse and specular albedos, outside the span of 3DMM, with faithful separation between them avoiding baking residual self-shadows in the albedo.
More recently, \cite{tran2018nonlinear, tran2019towards} use supervised training to learn a non-linear 3DMM model to produce more detailed albedo and geometry. \cite{saito2017photorealistic,laine2017production,bagautdinov2018modeling} show vast improvements in geometry reconstruction. \cite{huynh2018mesoscopic} improves on these approaches further by inferring mesoscopic facial attributes given monocular facial images, an attribute we do not model in our final reconstruction.
Finally, \cite{yamaguchi2018high} and \cite{lattas2020avatarme} use image-to-image translation networks to directly regress diffuse and specular albedos but they do not model light. Self-shadows can be observed in the albedo they obtain, whereas we model self-shadows implicitly. Additionally, these methods require ground-truth data to train the generative model, which is not easy to acquire. 
\section{Method}
\label{sec:method}
\begin{figure*}
\includegraphics[width=\linewidth]{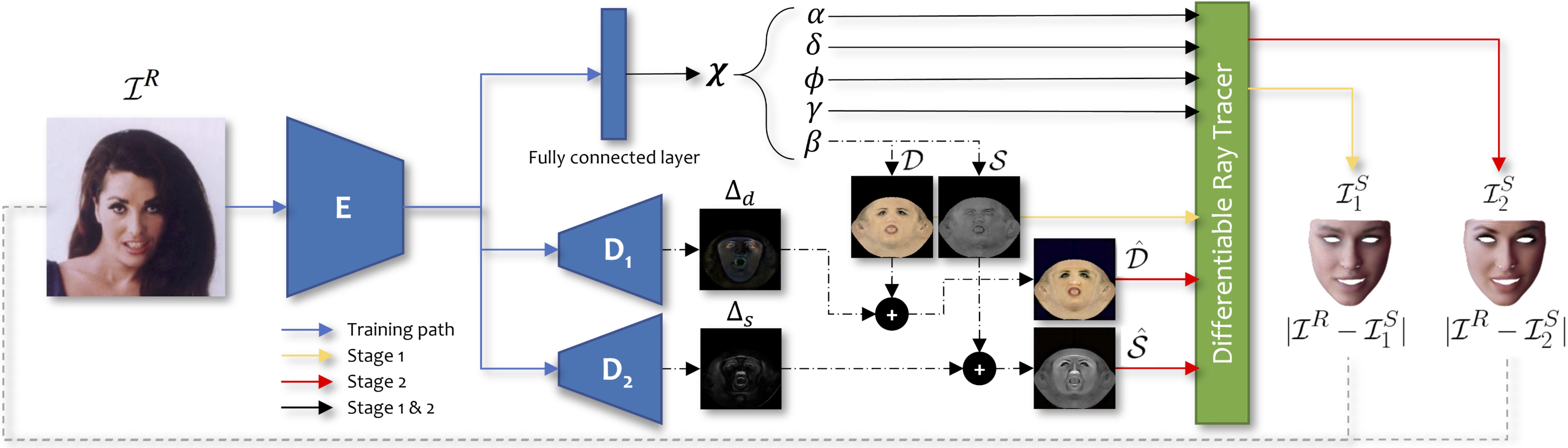}
  \caption{Method overview: A network $\mathbf{E}$ regresses semantic face attributes (shape, expression, light, statistical diffuse and specular albedos and camera). Two additional networks $\mathbf{D_1, D_2}$ are used to estimate increments $\Delta_d, \Delta_s$ on top of the statistical albedos to capture personalized reflectance (diffuse and specular) outside of the statistical prior space. A differentiable ray tracer is used for image formation.} 
  \label{fig:overview}
  \vspace{-5px}
\end{figure*}
\ad{Our method is inspired by the work of \cite{dib2021practical}, which achieves face reconstruction with personalized albedos outside the statistical albedo prior space and in challenging lighting conditions using ray-tracing within an optimization framework. While their method achieves state-of-the art results, it lacks robustness against initial starting point and is inherently slow, which makes their method not suitable for in-the-wild conditions. Our method overcomes these limitations by leveraging the generalization capacity of self-supervised learning.}

Our method is composed of two stages as shown in Figure \ref{fig:overview}. In stage I, an input image $\mathcal{I}^R$ is passed through a deep network $\mathbf{E}$\footnote{For the sake of conciseness, in the rest of the paper, we use E to denote the combination of this encoder with the fully connected layer.} (a pre-trained ResNet-152) followed by a fully connected layer to predict a semantic attribute vector $\chi$  for 3DMM shape ($\alpha$), expression blendshapes ($\delta$), camera pose ($\phi = \{R, T \}$, composed of rotation and translation), light ($\gamma$) modeled with 9 spherical harmonics bands, albedo prior ($\beta$), which encodes diffuse and specular priors from \cite{smith2020morphable}. Statistical diffuse $\mathcal{D}$ and specular $\mathcal{S}$ textures are obtained from $\beta$. These parameters are fed to a differentiable ray tracer to generate a ray traced image $\mathcal{I}_1^S$. This encoder $\mathbf{E}$ is trained end-to-end in a non-supervised manner to obtain a `base' reconstruction. At this level, the estimated albedos only capture low-frequency skin attributes. In stage II, to enhance these albedo priors, we train two additional decoders, $\mathbf{D_1}$ and $\mathbf{D_2}$, in a self-supervised way to estimate diffuse $\Delta_d$ and specular $\Delta_s$ increments to be added on top of the previously estimated textures, $\mathcal{D}$ and $\mathcal{S}$, respectively. The resultant textures, $\mathcal{\hat{D}}$ and $\mathcal{\hat{S}}$, are used to generate a new image $\mathcal{I}_2^S$. \ad{The challenges is to avoid mixing the diffuse and specular parts and baking unexplained shadows (residual) in the personalized albedos.}
In the next section, we briefly describe the scene attributes used for image formation, and then we discuss the training of the different networks.
\subsection{Scene attributes}
\label{ssec:sceneAttributes}
\textbf{Geometry} We use \cite{blanz1999morphable, gerig2018morphable}'s statistical face model, where identity is given by $\mathsf{e} = \mathsf{a}_s + \mathsf\Sigma_s \mathsf{\alpha}$. $\mathsf{e}$ a vector of face geometry vertices  with $N$ vertices. The identity-shape space is spanned by $\mathsf\Sigma_s \in \mathbb{R}^{3N \times K_s}$ composed of $K_s = 80$ principal components of this space. $\mathsf{\alpha} \in \mathbb{R}^{K_s}$ describes weights for each coefficient of the 3DMM and $\mathsf{a}_s\in\mathbb{R}^{3N}$ is the average face mesh. We use linear blendshapes to model face expressions over the neutral identity $\mathsf{e}$. $\mathsf{v} = \mathsf{e} + \mathsf\Sigma_e \mathsf{\delta}$, where $\mathsf{v}$ is the final vertex position displaced from $\mathsf{e}$ by blendshape weights vector $\mathsf{\delta} \in \mathbb{R}^{ K_e}$ and $\mathsf\Sigma_e \in \mathbb{R}^{3N \times K_e}$ composed of $K_e = 75$ principal components of the expression space.
\vspace{4px}
\\
\textbf{Reflectance} 
A simplified Cook-Torrance BRDF~\cite{cook1982reflectance,walter2007Microfacet}, with a constant roughness term, is used to model skin reflectance. In contrast to Lambertian BRDF, the Cook-Torrance BRDF can model specular reflections, and thus defines for each vertex $\mathsf{v}_i$ a diffuse $\mathsf{c}_i \in \mathbb{R}^3$ and a specular $\mathsf{s}_i\in \mathbb{R}^3$ albedos.
The statistical diffuse albedo $\mathsf{c}\in\mathbb{R}^{3N}$ is derived from 3DMM as $\mathsf{c} = \mathsf{a}_r + \mathsf\Sigma_r \mathsf\beta$, where $\mathsf\Sigma_r\in\mathbb{R}^{3N \times K_r}$ defines the PCA diffuse reflectance with $K_r=80$ and $\mathsf\beta\in\mathbb{R}^{K_r}$ the coefficients. $\mathsf{a}_r$ is the average skin diffuse reflectance. Similarly \ad{to \cite{dib2021practical}}, we employ the statistical specular prior introduced by \cite{smith2020morphable} to model the specular reflectance: $\mathsf{s} = \mathsf{a}_b + \mathsf\Sigma_b \mathsf\beta$ where $\mathsf\Sigma_b\in\mathbb{R}^{3N \times K_r}$ defines the PCA specular reflectance. $\mathsf{a}_b$ is the average specular reflectance. 
We use the same coefficients $\beta$ to sample for the diffuse and specular albedos as suggested by \cite{smith2020morphable}.
In unwrapped (UV) image texture space, $\mathcal{D} \in \mathbb{R}^{M \times M \times 3}$ and $\mathcal{S} \in \mathbb{R}^{M \times M\times3}$ are the statistical diffuse and specular albedos,  with $M \times M$ being the texture resolution.
\vspace{4px}
\\
\textbf{Illumination} Specular reflections are much more sensitive to light direction than diffuse reflections. So while in the literature 3-order spherical harmonics have been widely used in conjunction with Lambertian BRDF, we resort to nine SH bands with the Cook-Torrance model. \CeT{We actually found SH to be better suited to our deep learning framework than an explicit spatial representation as the one used by \cite{dib2021practical}}.
We show in section \ref{sec:ablation} that high-order SH produces better shadows estimation than low-order SH.
To use with the ray tracer, an environment map of $64 \times 64$ is derived from this light representation. We define $\gamma \in \mathbb{R}^{9 \times 9 \times 3}$ as the light coefficients to be predicted by the network $\mathbf{E}$.
\vspace{4px}
\\
\textbf{Camera} We use the pinhole camera model with rotation $\mathsf{R} \in \mathsf{SO}(3)$  and translation $\mathsf{T} \in \mathbb{R}^3$. We define $\phi = \{ \mathsf{T}, \mathsf{R}\}$ the parameters predicted by $\mathbf{E}$.
\subsection{Training}
\label{ssec:trainingFirstStage}
\begin{figure*}[!htb]
\centering
\includegraphics[width=0.95\linewidth]{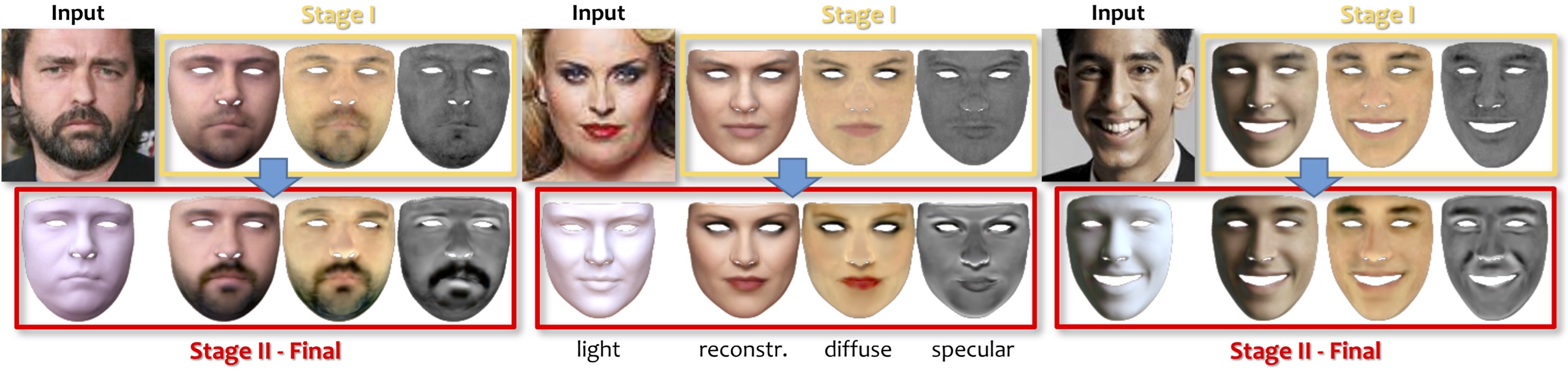}
  \caption{Yellow box: Base reconstruction (Stage I section \ref{sec:method}) with the estimated statistical albedo priors. Red box:  Final reconstruction (Stage II section \ref{sec:method}) with the final albedos and light.}
  \label{fig:results}
  \vspace{-5px}
\end{figure*}
\textbf{First stage (Base reconstruction)} In this stage, we use statistical geometry and albedo priors to obtain a first estimation of geometry, reflectance, light and camera attributes.
We define $\chi = \{\mathsf\alpha, \mathsf\delta, \mathsf\phi, \mathsf\gamma, \mathsf\beta\}$ as the semantic attribute vector used by the differentiable ray tracer to obtain a ray traced image $\mathcal{I}_1^S$. We train a deep encoder $\mathbf{E}$ to directly regress $\chi$. We use a pixel-wise photo-consistency loss between $\mathcal{I}_1^S$ and the input image $\mathcal{I}^R$:
\begin{equation}
 \label{eq:energy}
\mathsf{E}_{ph}(\chi) =\sum_{i\in\mathcal{I}} |\mathsf{p}_i^S(\chi) - \mathsf{p}_i^R|
\end{equation}
Here, $\mathsf{p}_i^S, \mathsf{p}_i^R \in\mathbb{R}^3$ are ray traced and real image pixel colors, respectively. Rendered pixel colors are given by $\mathsf{p}_i^S=\mathcal{F}(\chi)$, where $\mathcal{F}$ is the Monte Carlo estimator of the rendering equation \cite{Kajiya86Rendering}.
 We also define a sparse landmark loss $\mathsf{E}_{land}$, which measures the distance between the projection of $L = 68$ facial landmarks and their corresponding pixel projections $\mathsf{z}_l$ on the input image (\ad{more details in \cite{dib2021practical})}. These landmarks are obtained using an off-the-shelf landmarks detector \cite{bulat2017far}.
\md{$\mathsf{E}_{land}$ plays an important role as it helps stabilizing the training, and it mitigates the effect of the noisy gradients on the edges introduced by the ray tracer by guiding the training toward the correct shape around the edges -- especially mouth and eyes. } 
During training, we minimize the following energy function:
\vspace{-8px}
\begin{align}
\label{eq:loss1}
    \operatorname*{argmin}_{(\mathsf\alpha, \mathsf\delta, \mathsf\phi, \mathsf\gamma, \mathsf\beta)} \mathsf{E}_{d}(\chi) + \mathsf{E}_{p}(\mathsf\alpha, \mathsf\beta) + \mathsf{E}_{b}(\mathsf\delta)
\end{align}
$\mathsf{E}_{d}(\chi) = \mathsf{E}_{ph}(\chi) + \alpha_1\;\mathsf{E}_{land}(\chi)$ and $\mathsf{E}_{p}$ is the statistical face (shape and albedo) prior (\cite{dib2021practical}) that regularizes against implausible face geometry and reflectance deformations.
$\mathsf{E}_{b}(\mathsf\delta)$ is a soft-box constraint that restricts $\mathsf\delta$ to range $[0, 1]$.
\\
\textbf{Second stage (personalized albedos)}
In Figure \ref{fig:results} (yellow boxes) we show the final reconstruction together with the estimated statistical albedos ($D$ and $S$) obtained by the base reconstruction of stage I.
This result misses subject specific skin features, like beards and make-up, as 3DMM does not support them. We aim to personalize the albedo. However, the challenges are to avoid mixing the diffuse and specular parts and to avoid baking unexplained shadows (residual) in the personalized albedos. We train two additional networks $\mathbf{D_1}$ and $\mathbf{D_2}$, which take as input the latent space of $\mathbf{E}$ and estimate a diffuse and specular increments $\Delta_d$ and $\Delta_s$ which are added to $D$ and $S$, respectively. $\hat{D} = \mathcal{D} + \Delta_d$ and $\hat{S} = \mathcal{S} + \Delta_s$ are then used by the differentiable ray tracer to generate a new synthetic image $\mathcal{I}_2^S$. Predicting an increment on statistical texture priors instead of directly estimating a complete texture is important to force $\mathbf{E}$ to produce a good albedo prior while at the same time estimating a good increment over these priors. 
We define $\hat{\chi} = \{\mathsf\alpha, \mathsf\delta, \mathsf\phi, \mathsf\gamma, \hat{D}, \hat{S}\}$ which is used in the photo-consistency loss (eq \ref{eq:energy}). The following energy function is minimized:
\begin{align}
\label{eq:loss2}
\operatorname*{argmin}_{(\Delta_d, \Delta_s)} & \mathsf{E}_{d}(\hat{\chi}) + w_1( \mathsf{E}_{s}(\mathcal{\hat{D}}) + \mathsf{E}_{s}(\mathcal{\hat{S}})) + \nonumber\\ & w_{2D} \mathsf{E}_{c}(\mathcal{\hat{D}}, \mathcal{D}) + w_{2S} \mathsf{E}_{c}(\mathcal{\hat{S}}, \mathcal{S}) + \nonumber\\ &
 w_3( \mathsf{E}_{m}(\mathcal{\hat{D}}) +  \mathsf{E}_{m}(\mathcal{\hat{S}}))  + \nonumber\\ & w_4( \mathsf{E}_{b}(\mathcal{\hat{D}}) + \mathsf{E}_{b}(\mathcal{\hat{S}}))
\end{align}
 where $\mathsf{E}_{b}$ is the soft box constraint that restricts the albedos to remain in an acceptable range $[0, 1]$. $\mathsf{E}_{m}$ is a constraint term which ensures local smoothness at each vertex, with respect to its first ring neighbors in the UV space, and is given by $\mathsf{E}_{m}(\mathcal{\hat{A}}) = \sum_{\mathsf{x}_j  \in \mathcal{N}_{\mathsf{x}_i}} ||(\mathcal{\hat{A}}(\mathsf{x}_j) - \mathcal{\hat{A}}(\mathsf{x}_i)||_2^2$, where $\mathcal{N}_{\mathsf{x}_i}$ is 4-pixel neighborhood of pixel $\mathsf{x}_i$.

\CeT{We use regularization similar to \cite{tran2019towards, dib2021practical} to prevent residual shadows to leak into the albedos.} $\mathsf{E}_{s}(\mathcal{\hat{A}})= \sum_{i\in M}| \mathcal{\hat{A}}(\mathsf{x}_i)) - \texttt{flip}(\mathcal{\hat{A}}(\mathsf{x}_i)))|_1$ is a symmetry constraint, where $\texttt{flip}()$ is the \textit{horizontal flip} operator. $\mathsf{E}_{c}(\mathcal{\hat{A}}, \mathcal{A})$ is a consistency regularizer, which weakly regularizes $\mathcal{\hat{A}}$ with respect to the previously optimized albedo $\mathcal{A}$ based on the chromaticity $\kappa$
of each pixel in the texture given by $\mathsf{E}_{c}(\mathcal{\hat{A}}, \mathcal{A} ) = \sum_{i\in M} |\kappa(\mathcal{\hat{A}}(\mathsf{x}_i))-\kappa(\mathcal{A}(\mathsf{x}_i))|_1$. 
\begin{figure}[h]
\includegraphics[width=\linewidth]{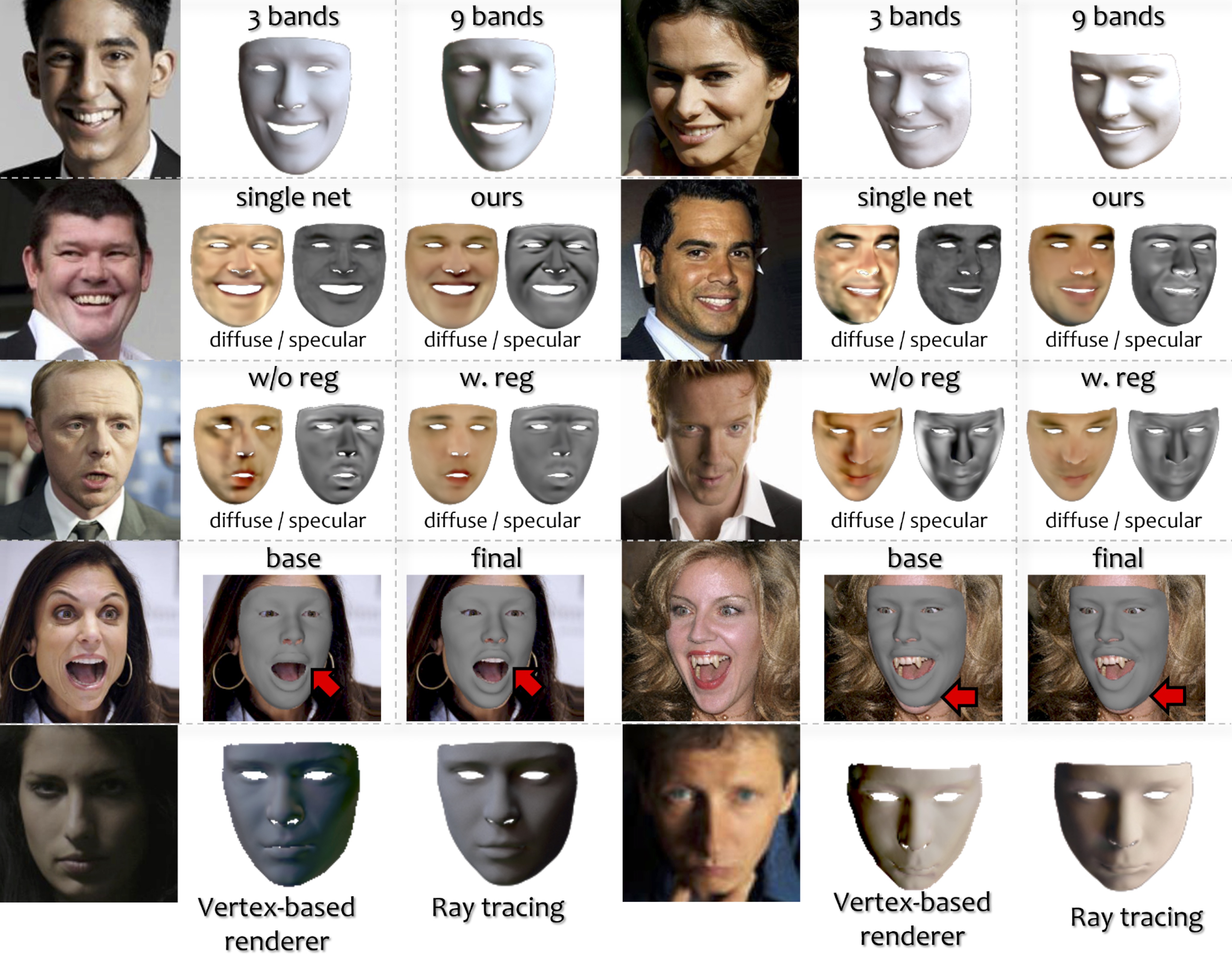}
  \caption{
  Row I: Comparison of the estimated light using 3 and 9 SH bands.
  Row II: Comparison of a single network vs. our dual-network approach to estimate albedo increments.
  Row III: Comparison of results obtained with and without the symmetry and consistency regularizers.
  Row IV: Comparison of base vs. final geometry reconstruction.
  Row V: comparison of vertex-based renderer and ray tracing.
  }
  \label{fig:ablationConsistencySecondStage}
  \vspace{-5px}
\end{figure}
\vspace{4px}
\\
\textbf{Training strategy}
Obtaining faithful and personalized diffuse and specular albedos (outside of the statistical prior space) in uncontrolled lighting conditions is an ill-posed problem. For instance, one can easily overfit the diffuse albedo increment $\Delta_d$ to explain the remaining information in the input image. To avoid this case, we proceed with the following strategy to separate the diffuse and specular albedos: After training $\mathbf{E}$ for few epochs, we fix $\mathbf{E}$ and start training $\mathbf{D_1}$ and $\mathbf{D_2}$ with a high regularization weight $w_{2D}$ for the diffuse consistency regularizer in order to keep $\mathcal{\hat{D}}$ closer to $D$. Next, we progressively relax the $w_{2D}$ constraint during the training to let the diffuse increment $\Delta_s$ capture more details. Finally, we train for all networks jointly ($\mathbf{E}$, $\mathbf{D_1}$ and $\mathbf{D_2}$). So, for a given image, we generate two images $\mathcal{I}_1^S$ and $\mathcal{I}_2^S$ and minimize the energy functions in \ref{eq:loss1} and \ref{eq:loss2} respectively and back-propagate over the whole attributes. This last step is important to stress $\mathbf{E}$ to produce better priors, light and pose while at the same time pushing $\mathbf{D_1}$ and $\mathbf{D_2}$ to capture more details in the albedos.
\begin{figure}[h]
\includegraphics[width=\linewidth]{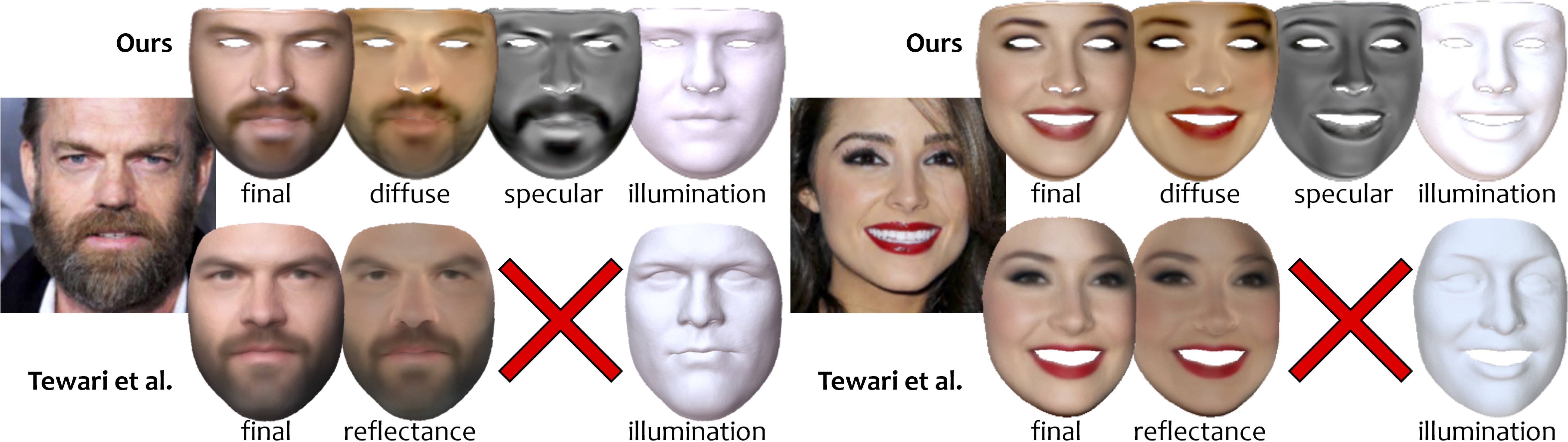}
  \caption{Comparison against \cite{tewari18fml} (subjects from authors paper).}
  \vspace{-10px}
  \label{fig:compMofa}
\end{figure}
\vspace{-5px}
\section{Results}
\label{sec:results}
For training, we used a total of $250K$ images, partially from CelebA dataset \cite{liu2015faceattributes}. For $\mathbf{E}$, we use a pre-trained ResNet-152, and both $\mathbf{D_1}$ and $\mathbf{D_2}$ networks use a cascade of 7 convolution layers. Ray tracing is based on the method of \cite{Li:2018:DMC}. Because ray tracing is very memory consuming, we use $256 \times 256$ as resolution for the texture and the input images. \ad{The inference takes $54$ ms ($47$ ms for $\mathbf{E}$ and $7$ ms for $\mathbf{D_1, D_2}$). During training, we use 8 samples per pixels for ray tracing the images.} Other implementation details can be found in the supplementary material (section I). Please note that images used in the figures were not used for training. The attributes estimated by our method are compatible with most rendering engines, nevertheless all results shown in the paper were rendered using ray tracing.
\\
Figure \ref{fig:results} (red boxes) shows reconstruction results from in-the-wild images for different subjects with various face attributes, head pose and lighting conditions. Our method successfully captures the facial hair and the lipstick -- outside of statistical albedo prior space -- for subjects on the left and in the middle respectively. Subject on the right is under challenging lighting conditions. Our method robustly estimates meaningful albedos and avoid baking residual shadows in the final albedos. We also show on Figure~\ref{fig:results} the effectiveness of albedo personalization provided by the networks $\mathbf{D_1}$ and $\mathbf{D_2}$ (red box) to refine the estimated statistical priors obtained by $\mathbf{E}$ (yellow box) from outside of the statistical albedo prior space. More results are shown on Figure \ref{fig:teaser} and in supplementary material (section 5).
\begin{figure}[h]
\includegraphics[width=\linewidth]{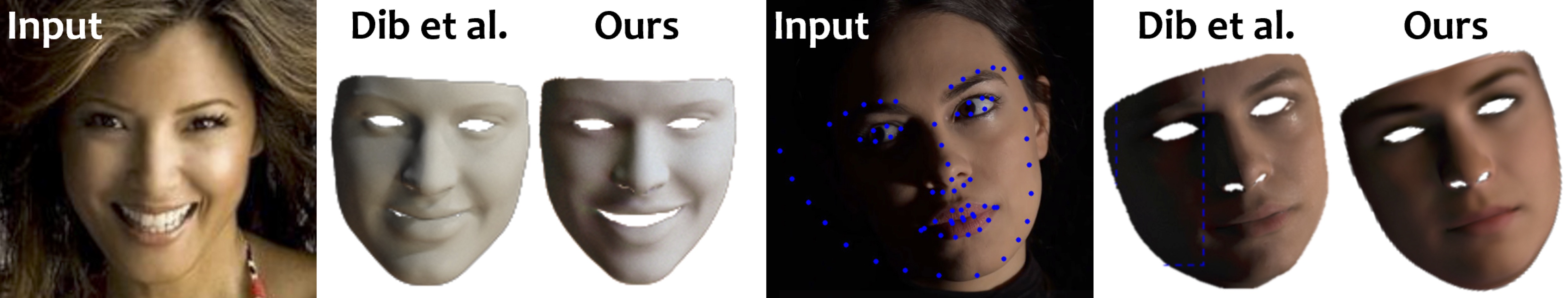}
  \caption{Comparison against \cite{dib2021practical} (right subject: authors paper).}
  \vspace{-15px}
  \label{fig:compDib}
\end{figure}
\section{Ablation}
\label{sec:ablation}
\textbf{Low-order SH parameterization}
In this experiment, we compare 9-order with 3-order SH. Estimated illuminations are shown in Figure~\ref{fig:ablationConsistencySecondStage} (first row). This shows that using higher-order spherical harmonics results in better shadow reconstruction, which prevents baking residual shadows in the albedo. 
In a similar experiment, 9-order SH has proven itself to be slightly better than 7-order.
\begin{figure*}
\includegraphics[width=\linewidth]{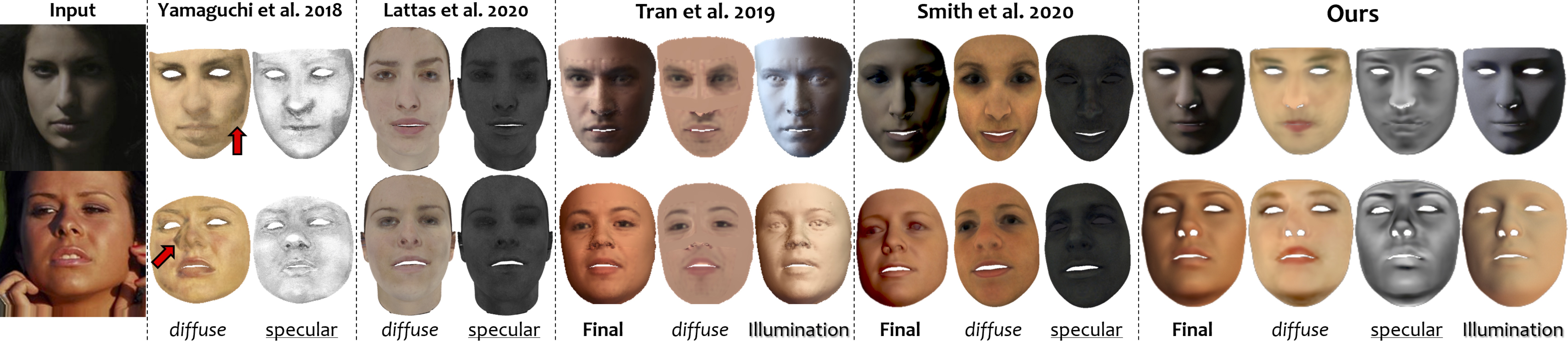}
  \caption{Comparison between methods - \cite{yamaguchi2018high}, \cite{lattas2020avatarme}, \cite{tran2019towards}, \cite{smith2020morphable}, and ours. The red arrows indicate the shadow baked in diffuse albedo.} 
  \label{fig:lattasYamagushiChenSmith}
  \vspace{-5px}
\end{figure*}
\vspace{4px}
\\
\textbf{Dual networks}
In this experiment, we trained a single network to regress diffuse $\Delta_d$ and specular $\Delta_s$ increments. As shown in Figure~\ref{fig:ablationConsistencySecondStage} (second row), a single network produces poor albedo separation compared to our approach that uses two separate networks, possibly because the diffuse and specular components capture different skin features that interfere when using a single network.
\vspace{4px}
\\
\textbf{Symmetry and consistency regularizers}
Figure~\ref{fig:ablationConsistencySecondStage} (row III) shows the importance of the regularizers used in equation \ref{eq:loss2}. In fact, because our light model may not always perfectly capture the real shadows, these regularizers prevent from baking residual shadows in the albedo.
\vspace{4px}
\\
\textbf{Joint training}
Training $\mathbf{E}$ jointly with $\mathbf{D_1}$ and $\mathbf{D_2}$ (refer to section \ref{sec:method}) improves the final geometry  as shown in  Figure~\ref{fig:ablationConsistencySecondStage} (row IV). The `final' mesh better fits the input image compared to the `base' mesh obtained from the base reconstruction (when training $\mathbf{E}$ only).
\vspace{4px}
\\
\textbf{Vertex-based renderer}
In this experiment, we train the same architecture, except that we used a `vertex-based' renderer, with the same illumination model (9-order SH) \cite{mahajan2007theory} and the simplified Cook-Torrance BRDF (more details in supp. material, section 2).
Figure~\ref{fig:ablationConsistencySecondStage} (last row) \fv{shows that ray tracing produces smoother and natural projected-shadows (especially around the nose).} 
\fv{This is because ray tracing can naturally models self-shadows while vertex-based renderer cannot. For instance, spherical harmonics (SH) coefficients are converted to an environment map (EM) for the use with ray tracing. Each pixel in EM acts as a light source at infinity. 
Shadows rays (rays shot from a surface point (P) towards a light source sampled from EM) are used to calculate a visibility mask for P. On the other hand, vertex-based renderers do not naturally model visibility of light sources. Work such as \cite{lyu2021efficient} tries to solve for this.}
\section{Comparison}
In the next, we compare, qualitatively and quantitatively, our method against  recent methods and to the ground-truth Digital Emily project and NoW benchmark \cite{RingNet:CVPR:2019}. 
\label{sec:comparison}
\vspace{4px}
\\
\textbf{Visual comparison} 
Figure \ref{fig:compMofa} shows comparison results of our method against \cite{tewari18fml} (results and images are taken from authors paper). Both methods achieve visually comparable results on the final reconstructed images. However, our method separates diffuse and specular albedos while \cite{tewari18fml} only estimates a reflectance map, which mixes diffuse and specular components. This makes our method more suitable for applications such as relighting.
\begin{figure*}
\includegraphics[width=\linewidth]{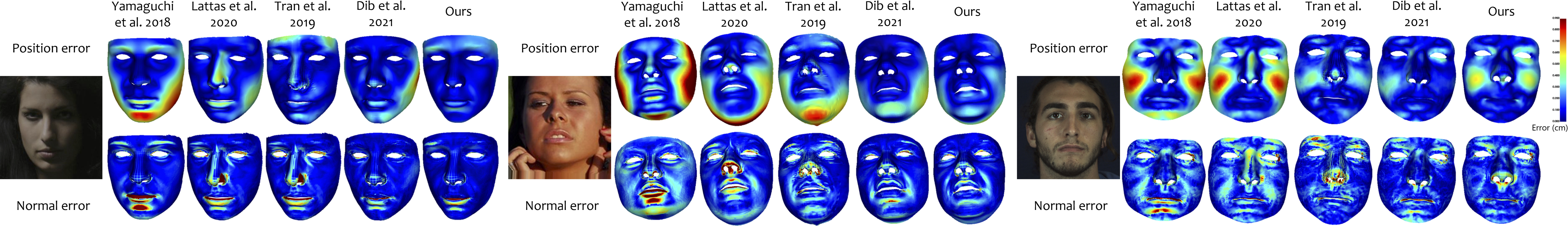}
  \caption{Vertex position and normal error for each method (from left \cite{yamaguchi2018high}, \cite{lattas2020avatarme}, \cite{tran2019towards}, \cite{dib2021practical}, and Ours) compared to GT mesh. 
  }
  \label{fig:compGeom}
  \vspace{-10px}
\end{figure*}

\ad{Figure \ref{fig:compDib} shows comparison against \cite{dib2021practical}. For the left subject, their method infers incorrect shape (around the mouth), head pose and illumination estimation. For the right subject, their estimated shape is inaccurate (right part of the head). This illustrates the sensitivity of \cite{dib2021practical} to the landmarks quality (as reported by the authors). Our method, completely independent of landmarks, is more robust to the lighting conditions and generates more convincing results.
In the supp. material (section 4), we show that our method achieves visually comparable results (\cite{dib2021practical} being slightly better) while being an order of magnitude faster ($54$ ms vs. $6.4$ min).
\fv{We note, that similarly to \cite{tewari2018HighFidelity}, combining both approaches by using our predicted attributes as initialization for the optimization produces better reconstruction quality.}
Finally, we note that \cite{dib2021practical} can capture shadows projected by point lights, while our method can only handle lights at infinite distance; nevertheless, our training strategy, together with carefully designed loss functions helps our method from baking unexplained shadows in the final personalized albedos at the expense of some albedo details (please refer to supp. material for more results in challenging lighting conditions).}

\begin{figure}[h]
\includegraphics[width=\linewidth]{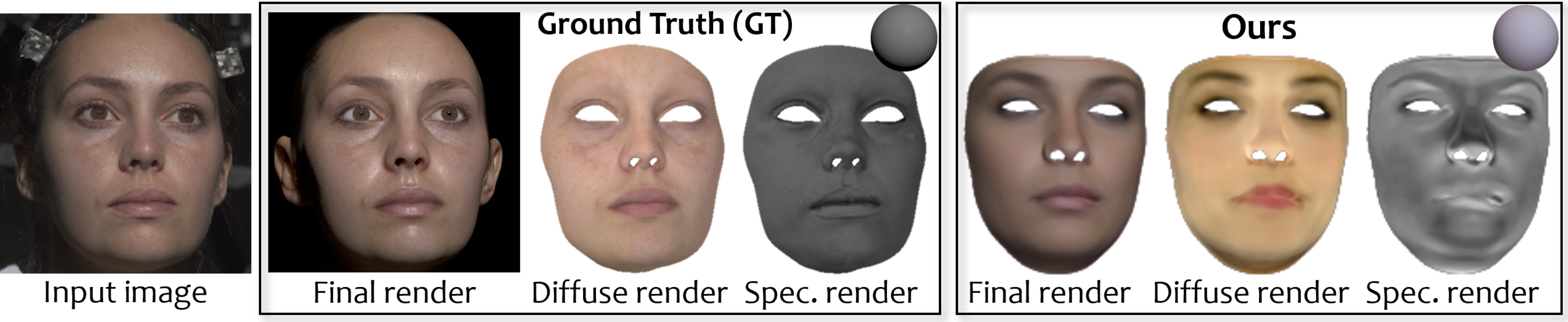}
  \caption{Our final, diffuse, and specular render images (Ours box) compared to GT (GT box). All GTs are rendered by Maya.}
  \label{fig:emilyEval}
  \vspace{-4px}
\end{figure}

Figure \ref{fig:lattasYamagushiChenSmith} shows comparison against \cite{yamaguchi2018high, lattas2020avatarme, tran2019towards, smith2020morphable}. Results for  
\cite{smith2020morphable} are obtained using their open-source implementations. For \cite{tran2019towards, yamaguchi2018high, lattas2020avatarme}, results are from original authors. \cite{yamaguchi2018high} and \cite{lattas2020avatarme} methods do not estimate light and thus final reconstruction is not available. 
Because \cite{yamaguchi2018high} does not model light, they bake some shadows in the diffuse albedo for both subjects. \ad{\cite{smith2020morphable} estimates light using 3-order SH but their reconstruction is bounded by the statistical albedo space and cannot capture personalized albedos outside of this space}. Compared to \cite{tran2019towards}, our method has `visually' better light estimation and smoother shadows (for first subject). Additionally, their diffuse and geometry have some artifacts visible around the nose. Also, their method does not estimate the specular component.
\cite{lattas2020avatarme} produces convincing shadow-free diffuse albedo. However, they do not estimate light and head pose.
\vspace{4px}
\\
\textbf{Geometric comparison}
We evaluate the quality of the reconstructed geometry on 23 images with 3D ground truth (GT) mesh from 3DFAW \cite{Pillai2019_3dfaw}, AFLW2000 fitting package \cite{zhu2017face}\cite{3ddfa_cleardusk}\cite{guo2020towards}, and from the wikihuman project\footnote{More details in supp. material section 3.} \cite{emilyWikihuman}. We compared our method to \cite{yamaguchi2018high, lattas2020avatarme, tran2019towards, dib2021practical} (Figure \ref{fig:compGeom} and Table \ref{tab:evalGeom}), using vertex position and normal errors calculated with respect to the GT meshes. Table \ref{tab:evalGeom} reports, for each method on all subjects, the average error ($\mu$) and standard deviation ($\sigma$) for vertex position and normal direction.
\aj{For vertex-position error, \cite{dib2021practical} scores the best average error ($0.174$), while \cite{tran2019towards} reports the second lowest measure. Comparing to the above methods, our method shows similar performance ($0.181$). We note that \cite{tran2019towards} learns a non-linear 3DMM model to improve the geometry while our method, which only uses 3DMM geometry, achieves comparable results. Our score is also on par with optimization-based method \cite{dib2021practical}, while being order of magnitude faster.  
For normal error, our method reports the second-best error ($0.148$), very close to \cite{dib2021practical}. Also, our score is better than \cite{tran2019towards} ($0.159$), since the mesh estimated by the latter has noise that appears around the nose (visible in Figure \ref{fig:lattasYamagushiChenSmith} and Figure \ref{fig:compGeom}).}
\fv{We also evaluate our method on the NoW benchmark \cite{RingNet:CVPR:2019} that only evaluates neutral mesh (no expression, albedo and pose evaluation). 
Nevertheless, we obtain very competitive results: 1.26/1.57/1.31mm (median/mean/std).}
\vspace{-3px}
\begin{table}
\caption{Vertex position and normal error $\mu$ and standard deviation $\sigma$ for each method on all subjects.}
\centering
\begin{adjustbox}{max width=0.8\linewidth}
\begin{tabular}{|l|ccccc|}
\hline
 & \cite{yamaguchi2018high} & \cite{lattas2020avatarme} & \cite{tran2019towards} & \cite{dib2021practical} & \textbf{Ours} \\
\hline
Position error $\mu$    (cm) & 0.236 & 0.184 & 0.176 & \textbf{0.174} & 0.181 \\
Position error $\sigma$ (cm) & 0.114 & 0.072 & 0.065 & 0.064 & 0.069 \\
\hline
Normal error $\mu$    (rad) & 0.152 & 0.156 & 0.159 & \textbf{0.139} & 0.148 \\
Normal error $\sigma$ (rad) & 0.051 & 0.043 & 0.046 & 0.046 & 0.048 \\
\hline
\end{tabular}
\end{adjustbox}
\label{tab:evalGeom}
\vspace{-5px}
\end{table}
\paragraph{Digital Emily}
As shown in Fig.\ref{fig:emilyEval}, we compare our method with Digital Emily \cite{emilyWikihuman} ground truth (GT). We note that our method bakes some albedo in the estimated light (as shown in the recovered environment map). 

We also compare quantitatively our image reconstruction quality against state-of-the-art (see Table~\ref{tab:emilyEval}). For each method, we compute SSIM \cite{zhou2004ssim} and PSNR scores versus GT, for final render, diffuse, and specular.
Since each method uses a different UV mapping, we compare the projection of the albedo on the input image (using the GT camera pose) and not on the unwrapped texture. For the `Final' rendered image, our method is on par with the method of  \cite{dib2021practical} and achieves better performance than \cite{smith2020morphable}. Since \cite{yamaguchi2018high} and \cite{lattas2020avatarme} do not estimate scene light, they do not have a final render image, so no comparison is available for these methods. For diffuse and specular albedos, \cite{yamaguchi2018high} is globally better than the other methods except for `Diffuse SSIM' and `Specular SSIM', where \cite{dib2021practical} and our method measure higher similarity, respectively.
\begin{table}[h]
\caption{Final render, diffuse and specular albedos in comparison with GT Maya renders. SSIM and PSNR (dB): higher is better.}
\centering
\begin{adjustbox}{max width=\linewidth}
\begin{tabular}{|c|cccccc|} 
\hline
{vs GT~} & {Final~} & {Final~} & {Diffuse~} & {Diffuse~} & {Spec.~} & {Spec.}  \\
{Render} & {(SSIM)} & {(PSNR)} & {(SSIM)}   & {(PSNR)}   & {(SSIM)} & ({PSNR)} \\
\hline
{Ours}                      & 0.933 & 35.932 & 0.653 & 29.548 & \textbf{0.642} & 29.274 \\
{\cite{dib2021practical}}   & \textbf{0.965} & \textbf{36.390} & \textbf{0.722} & 29.812 & 0.547 & 29.670 \\
{\cite{smith2020morphable}} & 0.906 & 35.389 & 0.639 & 29.006 & 0.452 & 28.833 \\
{\cite{lattas2020avatarme}} & -     & -      & 0.540 & 28.633 & 0.516 & 28.926 \\
{\cite{yamaguchi2018high}}  & -     & -      & 0.679 & \textbf{30.061} & 0.604 & \textbf{30.923} \\
\hline
\end{tabular}
\end{adjustbox}
\label{tab:emilyEval}
\vspace{-4px}
\end{table}
\\
%
%
%
\vspace{-17px}
\section{Limitations, Future works and Conclusion}
\label{sec:conclusion}
\begin{figure}[h]
\includegraphics[width=\linewidth]{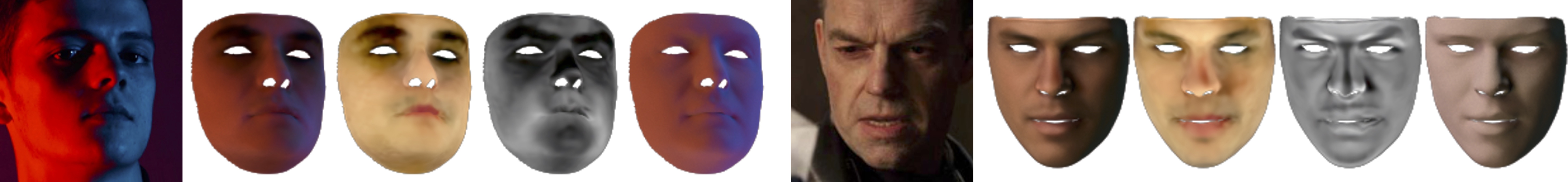}
  \caption{Limitations of our method.} 
  \label{fig:limitations}
  \vspace{-5px}
\end{figure}

\textbf{Limitations}
\ad{Our data-driven method inherits the bias of the training data, which does not provide high diversity in lighting and facial expression (most subjects in the dataset are with smiling faces, eyes opened and well lit). This leads to sub-optimal reconstruction for less frequent expression/lighting (incorrect light estimation for left subject in Figure \ref{fig:limitations}). 
This limitation can be mitigated by using a more balanced dataset while the method remains the same.} Also, with images with extreme shadows some artifacts may appear in the estimated albedos (right subject in Figure \ref{fig:limitations}). 

Disentangling light color from skin color from a single image is an ill-posed problem \fv{and is not solved in this work}. The limitations of 3DMM statistical albedo prior (unable to model all skin types e.g. non-Caucasian) make this separation even more difficult. 
Adding light color regularization, together with a more complete albedo prior can mitigate this.
\vspace{4px}
\\
\textbf{Future works} Our method could be extended by using a non-linear morphable model such as \cite{li2020learning} to improve the geometry at finer level. Using a more complex skin reflectance model such as BSSRDF/dielectric materials \cite{weyrich2006analysis} is also interesting.
Our method could also benefit from a work such as~\cite{zhang2020portrait}, which tackles the foreign (external) shadows challenge. Finally, our method can naturally extend to video-based reconstruction \cite{tewari2019fml}, which should improve the accuracy of the estimated facial attributes.
\\
\textbf{Conclusion}
In this work, we address the problem of face reconstruction under general illumination conditions, an important challenge to tackle for in-the-wild face reconstruction.
For this, we introduced the first deep-based and self-supervised method that achieves state-of-the art monocular face reconstruction in challenging lighting conditions. 
We build our work on recent methods, namely \cite{tewari17MoFA} which combines deep neural networks with differentiable rendering and \cite{dib2021practical} which uses ray tracing for face reconstruction within an optimization-based framework. Our method solves the limitations of \cite{tewari17MoFA} by using a better light and BRDF models and captures personalized diffuse and specular albedos outside of 3DMM space, while being robust against harsh shadows. 
Our approach also solves the limitations of \cite{dib2021practical} and achieves near-real time performance 
while at the same time being completely independent of landmarks at test time.
Our method naturally benefits from large-scale unlabeled data-sets. By comparing to recent approaches, we achieve better results in terms of robustness in scenes with challenging lighting conditions, while producing plausible reconstruction of subject-specific albedos. 
Beyond its robustness to lighting conditions, the rich reflectance decomposition produced by our method is compatible with existing rendering engines and allows for several style -- illumination and albedo -- transfer, edit applications, avatar creation and relighting.
\vspace{4px}
\\
\textbf{Acknowledgments.} This work has been supported by the ERC Consolidator Grant 4DReply (770784).
{\small
\bibliographystyle{ieee_fullname}
\bibliography{ms}
}
\clearpage
\appendix
\section{Implementation details}
\label{sec:impDetails}
We implemented the architecture using PyTorch~\cite{paszke2017automatic} with a GPU-enabled backend. Ray tracing is based on the method of \cite{Li:2018:DMC}, and for training we used Adam~\cite{kingma2014adam} as optimizer with default parameters. We used images from CelebA dataset \cite{liu2015faceattributes}  in addition to $40K$ images collected from the web for a total of $250K$ images. We keep $2K$ images for the validation. Images are aligned and cropped to a resolution of $256 \times 256$. We trained $\mathbf{E}$ for 10 epochs, then we fixed $\mathbf{E}$ and trained $\mathbf{D_1}$ and $\mathbf{D_2}$ for 5 epochs. Finally we trained all networks jointly for 5 epochs. We set our regularization weights as following: landmarks weight $\alpha_1 = 1$, $w_i = 0.002$, $w_c = 0.01$, symmetry regularizer $w_1 = 20$,  $w_{2S} = 0.01$, smoothness regularizer $w_3 = 0.0001$; and for $w_{2D}$, we start with $w_{2D} = 0.5$, and decrease it by a factor of $2$ at each epoch. For $\mathbf{E}$, we use a pre-trained \textit{ResNet-152} with latent space dimension equal to 1000.
Both $\mathbf{D_1}$ and $\mathbf{D_2}$ networks use a cascade of 7 convolution layers. Because ray tracing is very memory consuming, we use a texture resolution of $256 \times 256$ with batch size equal to $8$ and input image of resolution $256 \times 256$ to fit the GPU memory (12GB on a NVIDIA GeForce RTX 2080 Ti). For the learning rates, we use $1 e^{-6}$ for $\mathbf{E}$ and $1 e^{-7}$ for $\mathbf{D_1}$ and $\mathbf{D_2}$. For training, it takes 15 hours to do a single epoch. During training, we use 8 samples per pixels for ray tracing the images.
We experimented with different numbers of samples per pixel (spp) for ray tracing (8, 16 and 32 spp), but we did not obtain substantial improvements when using more than 8 spp, even though using 16 spp already made the training much slower. Additionally, as skin is generally not a highly specular surface, in our experiments, modeling self-geometry ray bounces did not lead to substantial gain in accuracy; thus we did not use it for training. The inference takes $54$ ms ($47$ ms for $\mathbf{E}$ and $7$ ms for $\mathbf{D_1, D_2}$).
\section{Vertex-based renderer implementation}
\label{sec:ablation}
\ad{In this section we provide implementation details of the vertex-based renderer that we used to compare against the ray tracer (please refer to section 5 in primary document). }

The vertex based renderer computes the irradiance by evaluating spherical harmonics (SH) for each vertex of the face mesh. To model skin reflectance, we use a simplified Cook-Torrance BRDF, thus the final irradiance is the sum of diffuse and specular irradiance terms. For the diffuse term, a spatial convolution with the \textit{half-cosine} is applied to the SH light representation. This corresponds to a multiplication of the SH coefficients ($B_{lm}$) of the light representation with SH coefficients ($A_{l}$) of the \textit{half-cosine} function (\cite{mahajan2007theory}). For each vertex, the diffuse irradiance, $\mathcal{B}_d$, is obtained by evaluating the resulting SH:
\begin{align}
    \mathcal{B}_d(\mathsf{n_i}, \mathsf{c_i}) = \mathsf{c_i} \cdot \sum_{l = 0}^{8} \sum_{m = -l}^{l} A_{l} \cdot B_{lm} \cdot Y_{lm}(\mathsf{n_i})
\end{align}
where $\mathsf{c}_i\in\mathbb{R}^3$ is the diffuse albedo of a vertex. $\mathsf{n}_i\in\mathbb{R}^3$ is the vertex normal.
The specular term is similarly obtained using a spatial convolution of the SH light representation with the BRDF kernel corresponding to the roughness (which is constant in the simplified Cook-Torrance BRDF model we use). The specular irradiance, $\mathcal{B}_s$, is obtained by evaluating the resulting SH:
\begin{align}
    \mathcal{B}_s(\mathsf{R_i}) = \sum_{l = 0}^{8} \sum_{m = -l}^{l} S_{l} \cdot B_{lm} \cdot Y_{lm}(\mathsf{R_i})
\end{align}
where $\mathsf{R_i}$ is the reflection direction of the viewing vector $\mathsf{W_i}$ according to the surface normal, and $S_{l}$ are the SH coefficients of the BRDF function corresponding to the roughness \cite{mahajan2007theory}.
The final irradiance $\mathcal{B}$ is equal to the sum of the diffuse and specular terms weighted by the specular intensity $s_i$:
\begin{align}
    \mathcal{B}(\mathsf{n_i}, \mathsf{c_i}, \mathsf{R_i}) = (1 - \mathsf{s_i}) \cdot  \mathcal{B}_d(\mathsf{n_i}, \mathsf{c_i})  + \mathsf{s_i} \cdot \mathcal{B}_s(\mathsf{R_i}) 
\end{align}
$\mathsf{s_i} \in\mathbb{R}$ is the specular albedo.

Finally, We use the following vertex-based photo-consistency loss to minimize during the training: 
\begin{equation}
 \label{eq:energy}
\mathsf{E}_{ph}(\chi) =\sum_{i = 1}^{N} |\mathcal{B}(\mathsf{n_i}, \mathsf{c_i}, \mathsf{R_i}) - \mathcal{I}^R ( \Pi \circ \mathsf{C(\mathsf{v}_i}))|
\end{equation}
where $N$ is the number of vertices,  $\mathsf{C(\mathsf{v_i}})$ is the projection of vertex $\mathsf{v_i}$ in the real image, equal to: $\mathsf{R}^{-1} (\mathsf{v_i} - \mathsf{T})$. $\Pi$ is the perspective camera matrix that maps a 3D vertex to a 2D pixel.
\section{Mesh difference}
\label{sec:meshDiff}
In this section, we provide more details on how the geometric error is calculated for each method (please refer to Table 1 in primary document).

The mean difference error is computed per-vertex on the entire mesh. We implement a 3D mesh evaluation protocol similar to \cite{Pillai2019_3dfaw}. For computing the mesh difference, we first align a reconstructed mesh towards a ground truth (GT) mesh. Several feature points, namely, sparse correspondence points are defined on both the GT and reconstructed facial meshes, where vertices are minimally affected by the facial muscles. With the corresponding points ready on both meshes, we use a traditional least-square estimation introduced by \cite{Umeyama1991LeastSquaresEO} to align the two meshes. After this alignment, we compute the distance from each vertex of a mesh to the other via a fast ray-triangle intersection method \cite{Moller1997inter}. The average error is computed for final difference between two meshes.

\section{More comparison results}
\label{sec:comp}
\begin{figure*}
\centering
\includegraphics[width=0.9\linewidth]{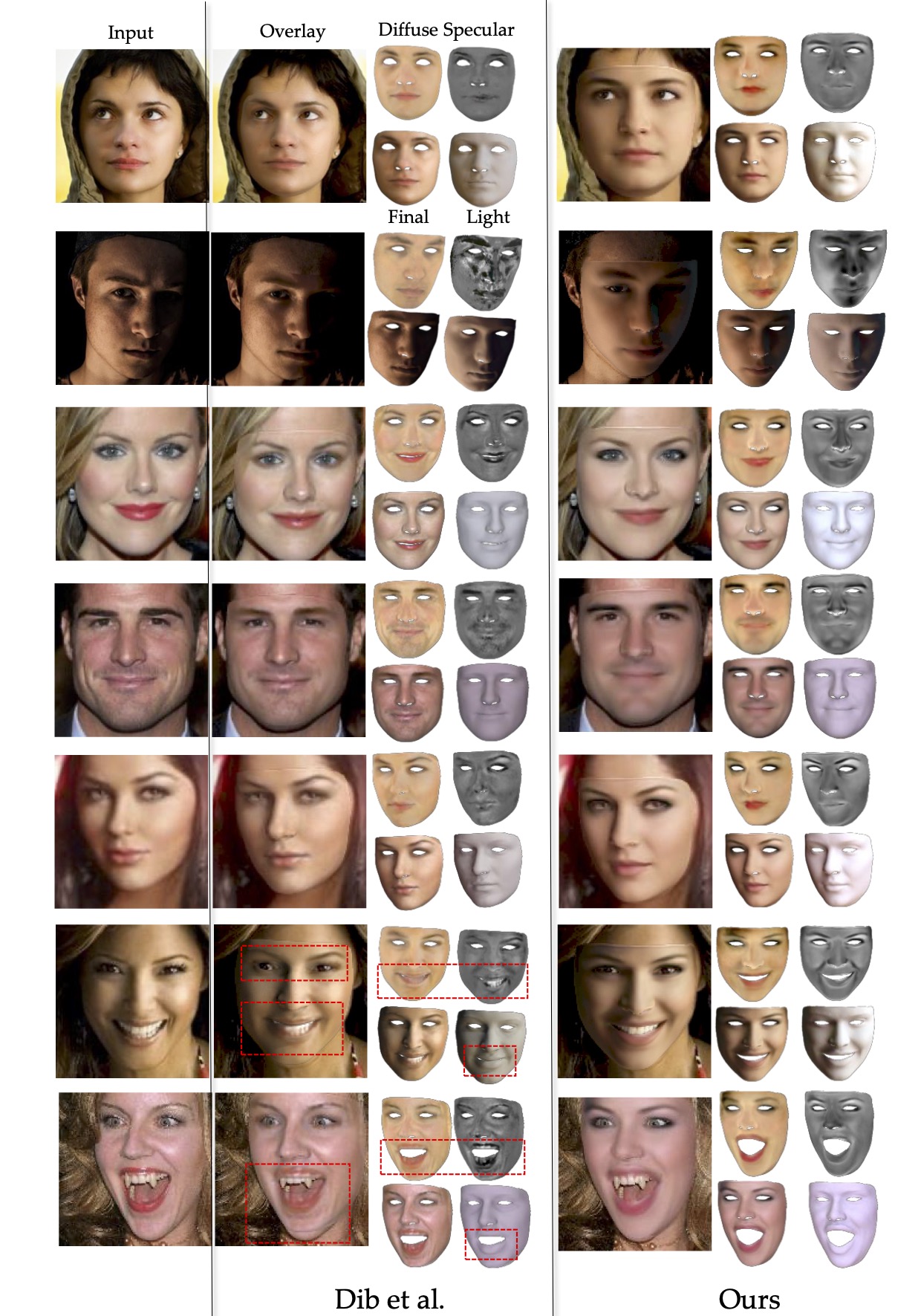}
  \caption{Comparison against \cite{dib2021practical}}
  \label{fig:compEG}
\end{figure*}

Figure \ref{fig:compEG} shows comparison results against the method of \cite{dib2021practical}. For each subject, we show the final reconstruction, estimated diffuse, specular and light for each method. The first two subjects are from the authors of \cite{dib2021practical}.
\\
Quite logically, the iterative optimization-based method of \cite{dib2021practical} achieves slightly better reconstruction results and captures more details in the estimated albedos. This is because \cite{dib2021practical} estimates and fine-tunes the facial and scene parameters specifically for each subject, while our method infers them directly without fine-tuning.
Nevertheless, our method is almost on par with \cite{dib2021practical} and can successfully handle some cases where \cite{dib2021practical} falters. For instance, with the last two subjects of the Figure \ref{fig:compEG}, in presence of shadows and strong expression, landmarks detector deliver less accurate initial starting points for the method of \cite{dib2021practical} which consequently gets trapped in wrong local minima.
This yields poor shape and artefacts in the estimated albedos  (highlighted in red boxes). Our method does not suffer from this limitation, proves to be more robust and produces visually more plausible reconstruction. We note that \cite{dib2021practical} estimates a roughness map, a parameter that we do not estimate. However, as reported by the authors, the missing statistical prior of the estimated roughness may sometimes yield to an over-fitting on this parameter.

\begin{figure*}
\includegraphics[width=\linewidth]{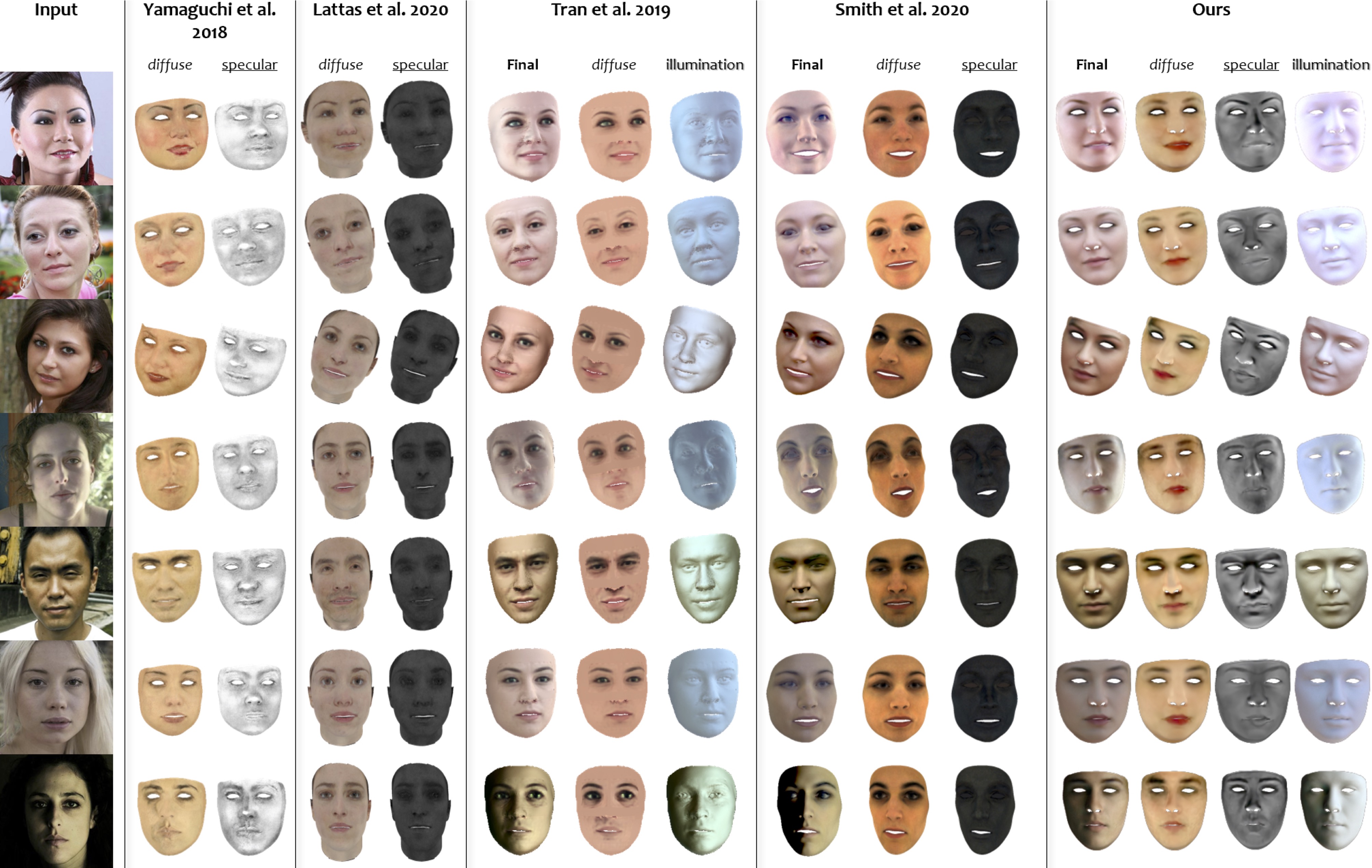}
  \caption{More visual comparisons against state-of-the-art methods.}
  \label{fig:moreCompRes}
\end{figure*}

\begin{figure*}
\includegraphics[width=\linewidth]{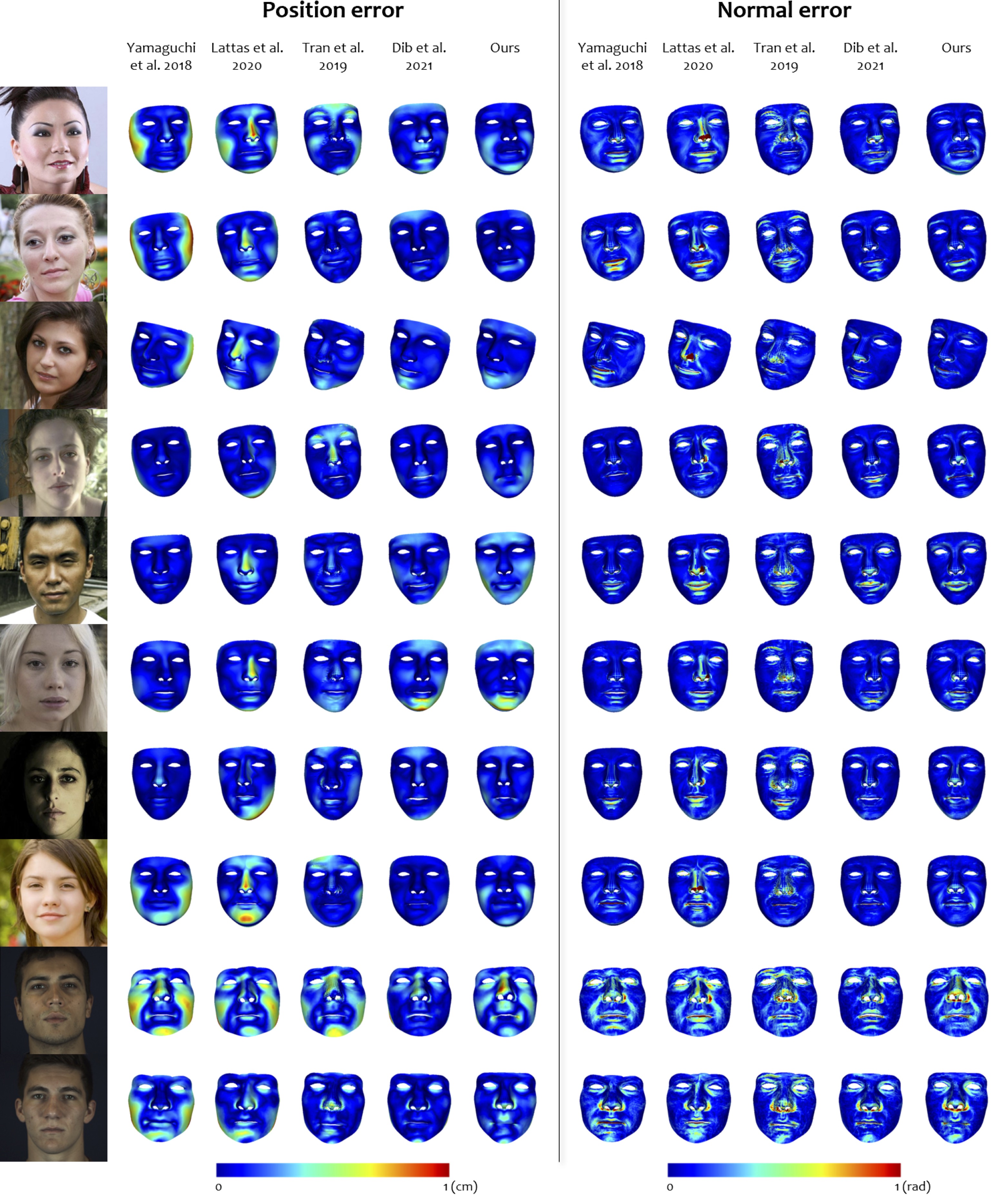}
  \caption{More geometric comparisons against state-of-the-art methods.}
  \label{fig:moreCompGeom}
\end{figure*}

We show more qualitative comparison against \cite{yamaguchi2018high}, \cite{lattas2020avatarme}, \cite{tran2019towards}, and \cite{smith2020morphable} in Figure \ref{fig:moreCompRes}. In Figure \ref{fig:moreCompGeom}, we show more quantitative comparison against \cite{yamaguchi2018high}, \cite{lattas2020avatarme}, \cite{tran2019towards}, and \cite{dib2021practical}.
\\

\section{Face catalog}
\label{sec:faceCatalog}
\begin{figure*}
\centering
\includegraphics[width=\textwidth,height=\textheight,keepaspectratio]{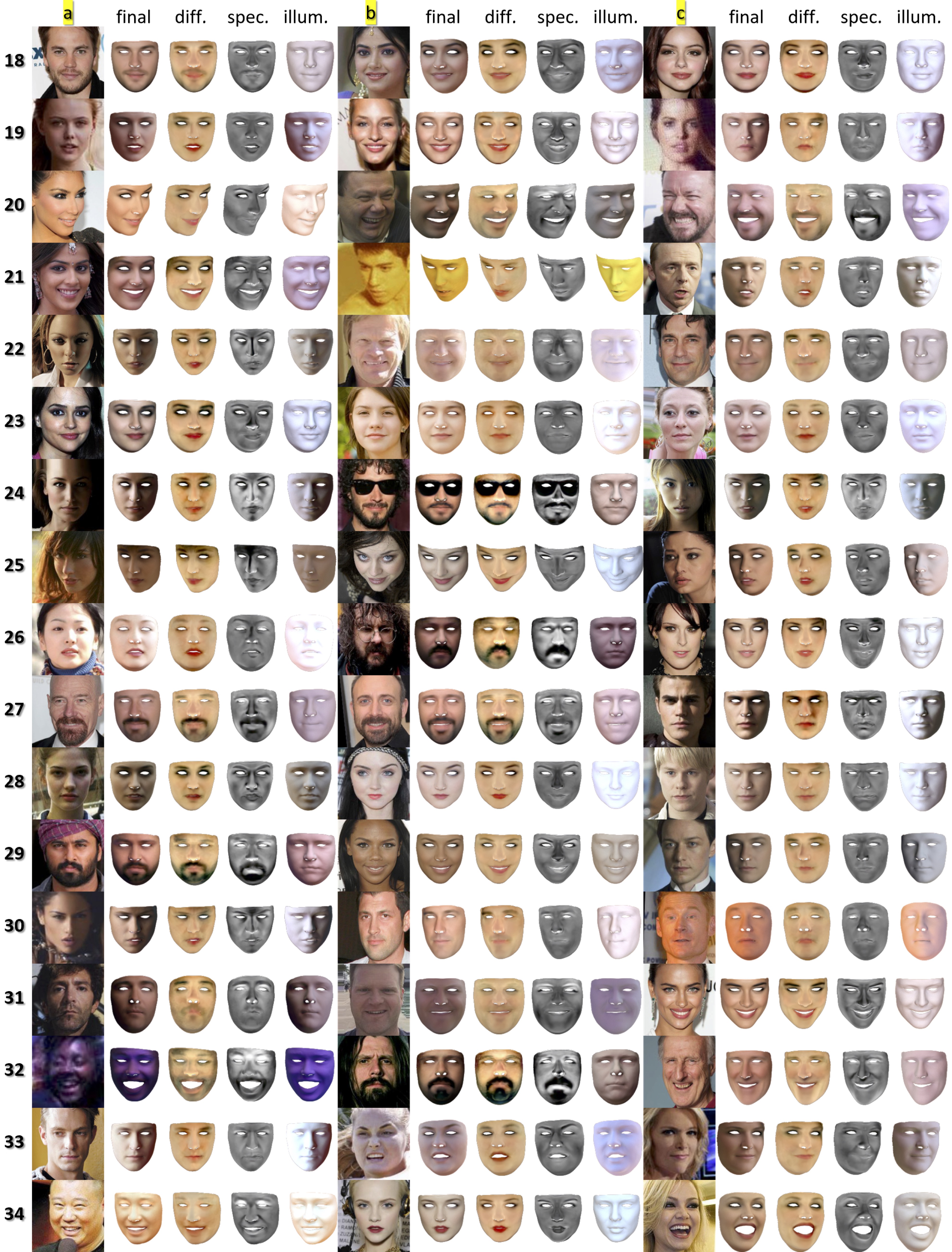}
\caption{Face catalog of our reconstruction. For each subject, we show the input, final, diffuse, specular, and illumination. More results are in the accompanied  video.}
\label{fig:faceCatalog}
\end{figure*}

In Figure \ref{fig:faceCatalog}, we show more reconstruction results from in-the-wild images. For each subject we show the final reconstruction and the estimated diffuse, specular albedos and illumination. More results are in the accompanied video.
\begin{figure*}
\centering
\includegraphics[width=\textwidth,height=\textheight,keepaspectratio]{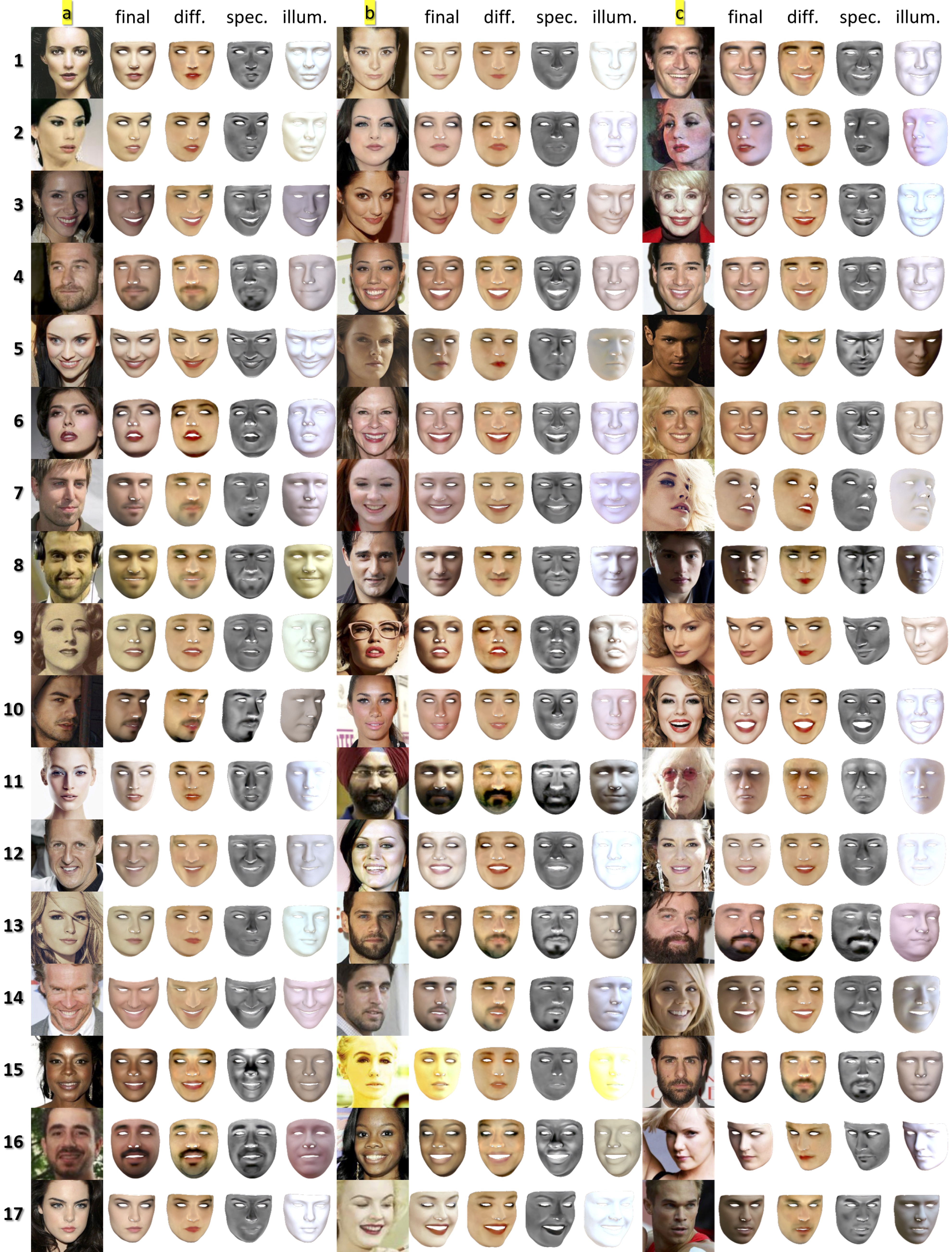}
\end{figure*}

\section{More relighting examples}
\label{sec:relighting}
\begin{figure*}
\centering
\includegraphics[width=\textwidth,height=\textheight,keepaspectratio, page = 1]{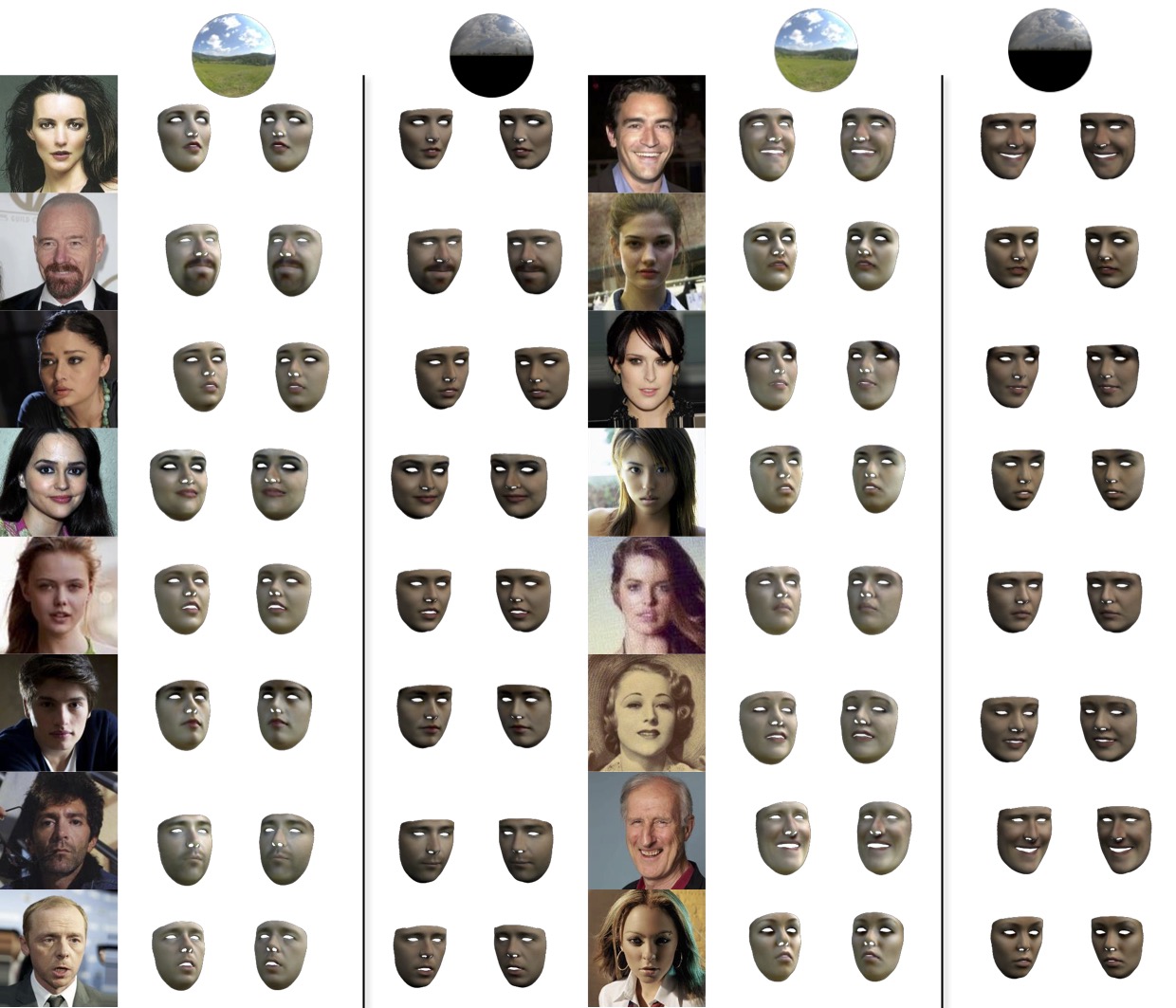}
\caption{Relighting examples (More relighting results are in the accompanied video).}
\label{fig:relighting1}
\end{figure*}

\begin{figure*}
\centering
\includegraphics[width=\textwidth,height=\textheight,keepaspectratio, page = 2]{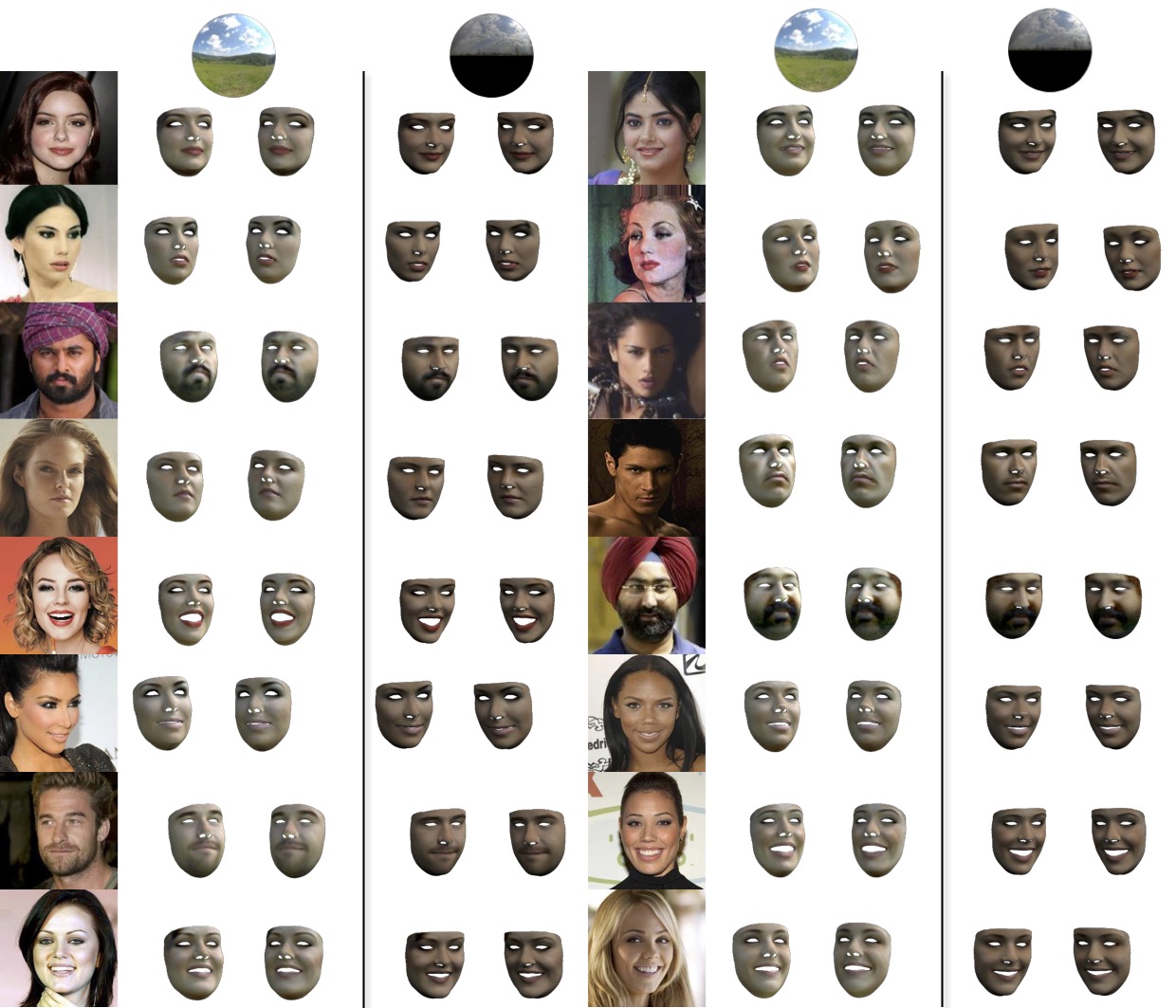}
\caption{Relighting examples (More relighting results are in the accompanied video). }
\label{fig:relighting2}
\end{figure*}

Figure \ref{fig:relighting1} and \ref{fig:relighting2} show more relighting examples where the estimated illumination is replaced with an environment-map.
More relighting results are in the accompanied video.

\end{document}